\pdfoutput=1

\documentclass[11pt]{article}
\usepackage[final]{acl}
\usepackage{times}
\usepackage{latexsym}

\usepackage[T1]{fontenc}

\usepackage[utf8]{inputenc}

\usepackage{microtype}

\usepackage{inconsolata}

\usepackage{graphicx}

\def \thetav{\boldsymbol{\theta}}

\usepackage{amssymb} 
\usepackage{graphicx}

\usepackage{graphicx}
\usepackage{algorithmicx,algorithm}
\usepackage[noend]{algpseudocode}
\usepackage{algorithmicx,algorithm}
\usepackage{bm}
\usepackage{bbm}
\usepackage{booktabs}       
\usepackage{amsmath}
\usepackage{amsfonts}
\usepackage{array}
\usepackage{bm}
\usepackage{multirow}
\usepackage{subcaption}
\usepackage{hyperref}       
\usepackage{url}            
\usepackage{graphicx} 
\usepackage{utfsym}
\usepackage{xcolor}
\usepackage{inconsolata}
\usepackage{CJKutf8}
\usepackage{pythonhighlight} 
\usepackage{makecell}
\usepackage[normalem]{ulem}

\title{CoViPAL: Layer-wise Contextualized Visual Token Pruning \\ for Large Vision-Language Models}

\author{
    Zicong Tang$^{2,\dagger}$, 
    Ziyang Ma$^{2}$, 
    Suqing Wang$^{2}$,
    Zuchao Li$^{1,\dagger}$\thanks{$\ $  Corresponding author. $^\dag$ Equal contribution. This work was supported by the National Natural Science Foundation of China (No. 62306216) and the Technology Innovation Program of Hubei Province (No. 2024BAB043).}, 
    Lefei Zhang$^{2}$, 
    Hai Zhao$^{3}$, \\ 
    \textbf{Yun Li}$^{4}$ \textbf{and} \textbf{Qianren Wang}$^{4}$\\
    {$^{1}$School of Artificial Intelligence, Wuhan University} \\
    {$^{2}$School of Computer Science, Wuhan University} \\
    {$^{3}$School of Computer Science, Shanghai Jiao Tong University} \\
    {$^{4}$Cognitive AI Lab, Shanghai Huawei Technologies, China} \\
    {\tt \{tangzc,maziyang,wangsuqing,zcli-charlie,zhanglefei\}@whu.edu.cn} \\
    {\tt zhaohai@cs.sjtu.edu.cn,lychina@139.com,wangqr2019@qq.com}
}

\begin{document}
\maketitle

\begin{abstract}
Large Vision-Language Models (LVLMs) process multimodal inputs consisting of text tokens and vision tokens extracted from images or videos. Due to the rich visual information, a single image can generate thousands of vision tokens, leading to high computational costs during the prefilling stage and significant memory overhead during decoding. Existing methods attempt to prune redundant vision tokens, revealing substantial redundancy in visual representations. However, these methods often struggle in shallow layers due to the lack of sufficient contextual information. We argue that many visual tokens are inherently redundant even in shallow layers and can be safely and effectively pruned with appropriate contextual signals. In this work, we propose CoViPAL, a layer-wise contextualized visual token pruning method that employs a Plug-and-Play Pruning Module (PPM) to predict and remove redundant vision tokens before they are processed by the LVLM. The PPM is lightweight, model-agnostic, and operates independently of the LVLM architecture, ensuring seamless integration with various models. Extensive experiments on multiple benchmarks demonstrate that CoViPAL outperforms training-free pruning methods under equal token budgets and surpasses training-based methods with comparable supervision. CoViPAL offers a scalable and efficient solution to improve inference efficiency in LVLMs without compromising accuracy. Our code is available in \url{https://github.com/tyxqc/CoViPAL}.
\end{abstract}

\begin{figure}
    \centering
    \includegraphics[width=1\linewidth]{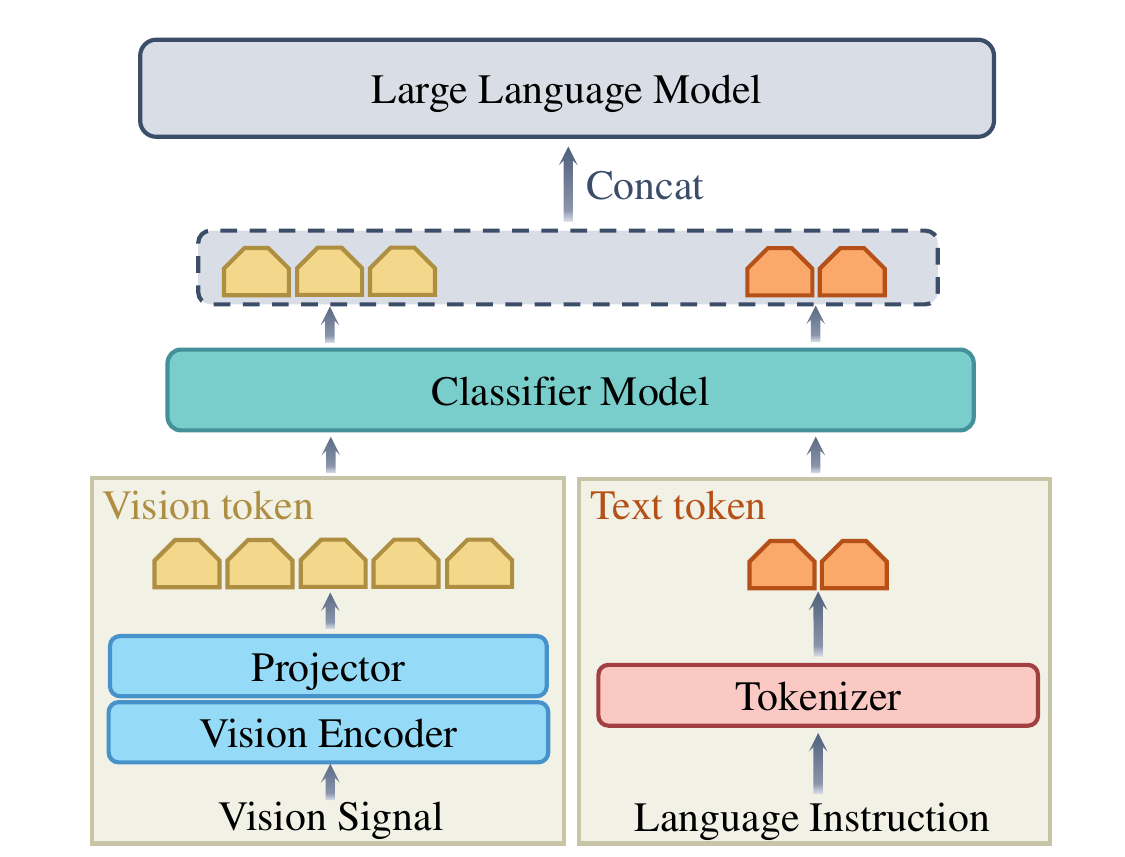}
    \caption{Illustration for CoViPAL at inference stage.}
    \label{fig:inference}
\end{figure}

\begin{figure*}[ht]
    \centering
    \begin{subfigure}[b]{0.44\linewidth}
        \centering
        \includegraphics[width=\linewidth]{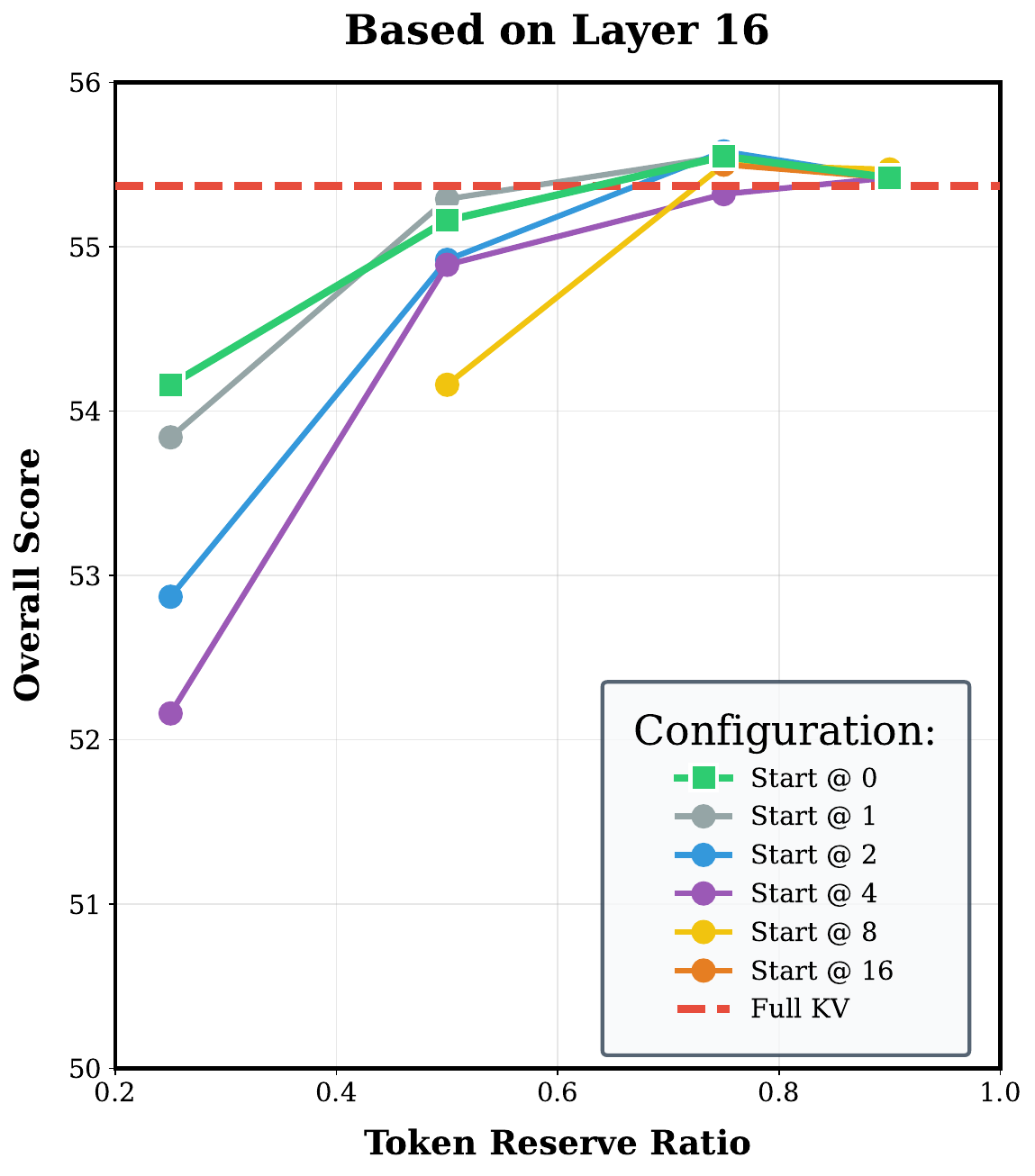}
        \caption{Pruning based on attention of layer 16. }
        \label{fig:observation_GQA}
    \end{subfigure}
    \hspace{0.01\linewidth}
    \begin{subfigure}[b]{0.46\linewidth}
        \centering
        \includegraphics[width=\linewidth]{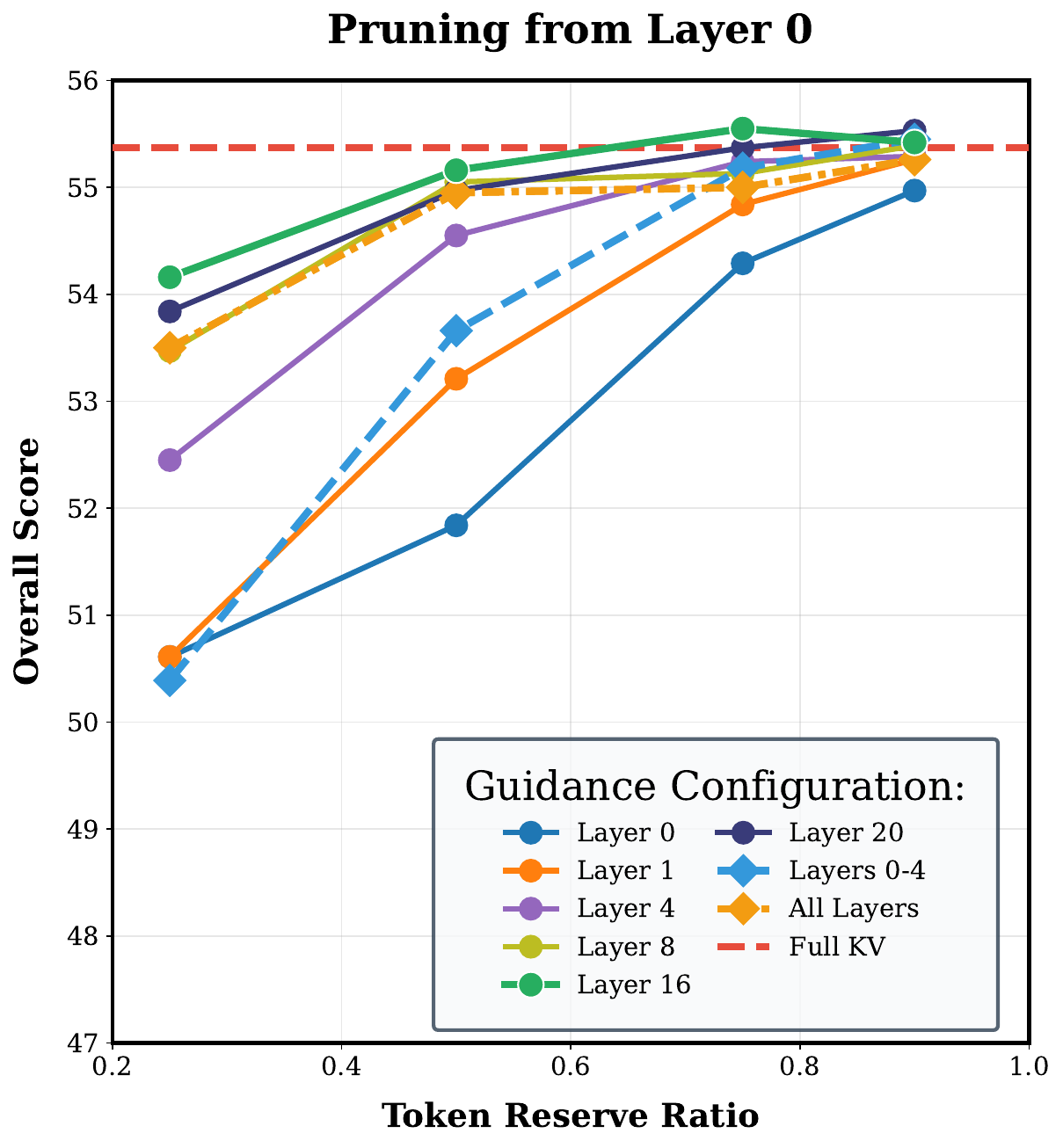}
        \caption{Pruning guided by different attention weight.}
        \label{fig:observation_MVbench}
    \end{subfigure}
    \caption{Prune tokens in different layers and based on different attention weights.}
    \label{fig:Observation}
\end{figure*}

\section{Introduction}
Large Vision-Language Models (LVLMs, ~\citealp{vicuna2023, gemeni, qwenvl, BatGPTCharLie}) have recently demonstrated remarkable capabilities in understanding and generating content grounded in visual inputs, including both images and videos. To effectively capture the rich spatial and semantic details inherent in visual signals, these models often rely on generating hundreds or even thousands of visual tokens per image or video. For instance, LLaVA-OneVision~\cite{llava-onevision} explicitly allocates up to 7,290 visual tokens per image, leveraging a large corpus of high-quality images to maximize visual comprehension. 

Although dense visual token representations enhance the model's capacity to understand fine-grained visual content, they come at the cost of substantial computational and memory overhead~\cite{zhang2024sparsevlm}. This leads to reduced inference efficiency and makes it difficult to apply LVLMs in scenarios where resources are limited or real-time performance is required.

To address this issue, prior work has explored reducing the number of visual tokens or compressing their corresponding key-value (KV) cache~\cite{bolya2022token}, highlighting the substantial redundancy present in visual representations. Token eviction methods discard less informative tokens based on importance scores~\cite{chen2024image}, while token merging approaches group similar tokens and consolidate them to reduce token number~\cite{chen2024efficient}.  Empirical observations suggest that pruning visual tokens in shallow layers can significantly hurt performance and every visual token matters in these layers~\cite{PyramidDrop}. Despite their effectiveness to some extent, these methods largely fail to prune tokens in the shallow layers.

Visual token reduction is less effective in shallow layers, primarily because tokens in these layers interact with fewer transformer decoder layers, resulting in limited contextual information. This makes it challenging to identify unimportant tokens, leading to significant performance degradation when attempting to prune visual tokens at these stages~\citep{KCDLuohe}. However, we observe that some visual tokens are inherently redundant and can be effectively and safely pruned when guided by appropriate contextual information. Based on this insight, we propose CoViPAL, a contextualized visual token pruning method that operates across all layers, as illustrated in Figure ~\ref{fig:inference}. CoViPAL implements the PPM module using a small classifier trained on limited data to identify and remove less important tokens before they are passed to the base model of LVLM, thereby reducing the number of visual tokens while maintaining model performance.

We conducted experiments on two models: LLaVA-OneVision and LLaVA-Video. For LLaVA-OneVision, we trained the classifier using only 0.46\% of the pretraining dataset, while for LLaVA-Video, we extended its capabilities to handle video inputs using just 7.4\% of the video instruction-following dataset. Additionally, we performed extensive experiments on various image and video benchmarks. The results demonstrate that our method reduces the prefilling time by up to 60\% compared to the original model, with only minimal performance degradation when pruning 75\% of the visual tokens. Furthermore, our approach outperforms both training-free methods, FastV and SparseVLM, and training-based method PyramidDrop, with the same visual token budget.

\section{Related Works}

\subsection{Token Pruning} Token pruning methods aim to remove tokens with low attention or feature similarity after early or intermediate layers~\citep{chen2024image, lin2025boosting, PyramidDrop, spindlekv1}, or optimize pruning schedules using small inference batches to meet FLOPs budgets~\citep{ye2025fit, SirLLM}. These methods generally prioritize the preservation of early tokens to avoid information loss~\citep{ModelHemorrhage,YaoYao}.

\subsection{Token Merging} Alternatively, similarity-based merging techniques fuse redundant tokens either spatially or cross-modally to reduce token count while maintaining semantic integrity and accuracy~\citep{chen2024efficient, shi2023crossget, Zhaoyi}. These methods achieve compression without compromising downstream performance. They typically leave the tokens in shallow layers unmerged to maintain overall performance.

\subsection{Hybrid Methods} Recent methods combine pruning and merging by ranking tokens based on attention, pruning low-importance tokens, and merging redundant ones to recycle information~\citep{zhong2024aim, shang2024llava, zhang2024sparsevlm}. For instance, LOOK-M~\citep{wan2024look} addresses long-context inference by compressing the KV cache through text-guided merging of similar key-value pairs, thereby reducing memory usage and improving decoding speed~\citep{kvlatent}.

These approaches generally retain visual tokens in shallow layers to minimize significant performance degradation. In contrast, our method demonstrates that visual token redundancy exists across all layers and can be safely pruned using a lightweight classifier trained on a small dataset. This approach facilitates earlier and more efficient pruning without sacrificing critical information.

\begin{figure*}[ht]
    \centering
    \begin{subfigure}[b]{0.46\linewidth}
        \centering
        \includegraphics[width=\linewidth]{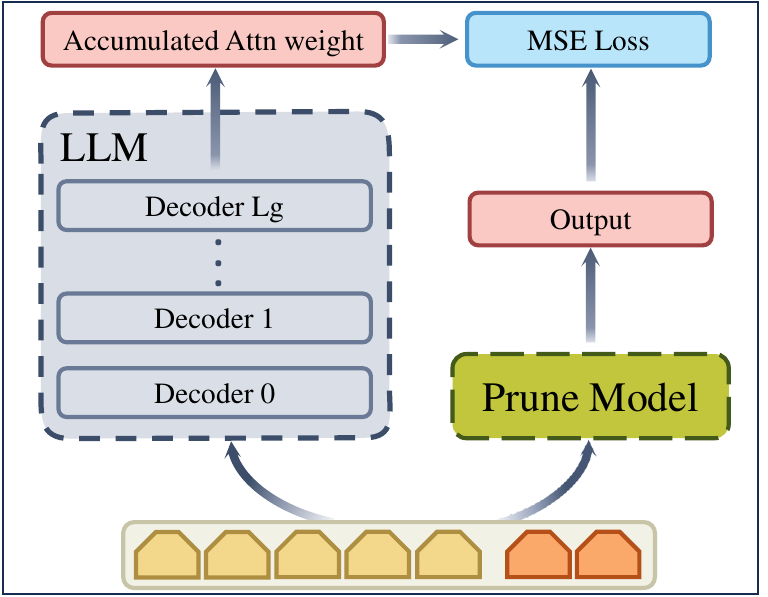}
        \caption{Stage 1: attention guided training. }
        \label{fig:Train1}
    \end{subfigure}
    \hfill
    \begin{subfigure}[b]{0.48\linewidth}
        \centering
        \includegraphics[width=\linewidth]{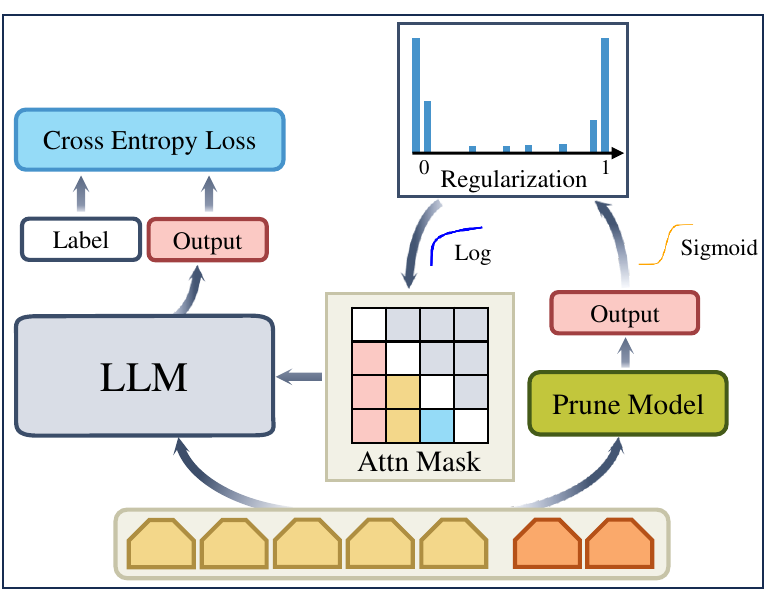}
        \caption{Stage 2: end-to-end training.}
        \label{fig:Train2}
    \end{subfigure}
    \caption{Two stage training strategy.}
    \label{fig:TrainStrategy}
\end{figure*}

\section{Preliminary}
\subsection{Notations}
In LVLMs, a vision encoder is typically employed to extract visual features, while a projector is used to map these features into the word embedding space. We denote the vision encoder and projector as \( g(\cdot) \), so the visual tokens are represented as \( \mathbf{H}_{\texttt{v}} = g(\mathbf{X}_{\texttt{v}}) \), where \( \mathbf{X}_{\texttt{v}} \) is the visual input. The textual input is represented by the text tokens \( \mathbf{H}_{\texttt{t}} \), which are concatenated with the visual tokens, forming the input to the LLM as \( f(\cdot) \).

For token pruning, we assign an importance score \( \mathbb{S} \) to each visual token. This score serves as the guiding criterion for the pruning process, directly determining the relevance of each token. Based on this score, we select the most important tokens to retain, while pruning those deemed less relevant, thereby reducing the overall number of visual tokens in the input. 
\subsection{Preliminary Experiment}
We conduct a preliminary study using LLaVA-OneVision-7b-Chat~\cite{llava-onevision} on the MVBench dataset~\cite{li2024mvbench}, where token pruning is applied at decoder layer \( L_p \), guided by attention weights from an earlier layer \( L_g \).

As shown in Figure~\ref{fig:Observation}, the choice of guidance layer \( L_g \) has a stronger impact on pruning effectiveness than the pruning layer \( L_p \) itself. This underscores the importance of selecting a semantically rich guidance layer. In particular, the 16th layer in LLaVA-OneVision proves to be a strong candidate for generating token importance scores.

Prior work often assumes \( L_p = L_g \), attributing pruning performance to the pruning layer rather than the quality of the guidance~\cite{zhong2024aim, zhang2024sparsevlm, lin2025boosting}. Our results challenge this assumption, showing that such coupling may lead to suboptimal pruning.

We observe that many visual tokens are inherently redundant and can be pruned with minimal performance loss when guided effectively. However, using deeper layers for guidance (\( L_g > L_p \)) introduces a trade-off: the model must prefill up to \( L_g \) to compute attention scores \( A_g \), then reprocess from \( L_p \) after pruning. This two-step procedure adds significant inference overhead.
\begin{figure}[h]
    \centering
    \includegraphics[width=0.6\linewidth]{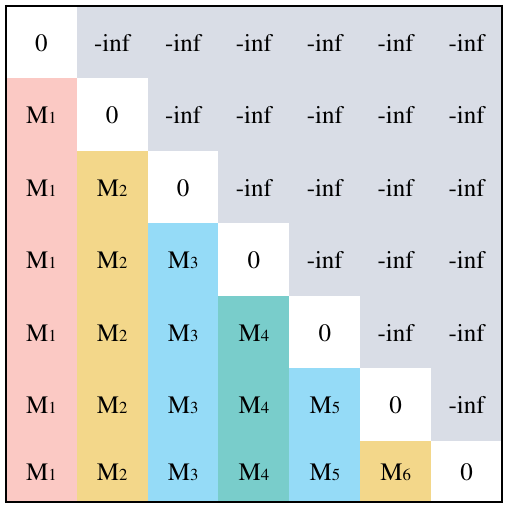}
    \caption{Attention mask at the training stage 2.}
    \label{fig:attn_mask}
\end{figure}
\section{Method}
\subsection{Inference}
Our observations indicate that some visual tokens are inherently redundant across layers, while the attention weights in shallow layers are not sufficiently effective at guiding the pruning. To address this, we employ a plug-and-play pruning classifier (referred to as the \textit{classifier}) to capture the inherent redundancy of the visual features for pruning.

We denote the classifier as \( p_{\theta}(\cdot) \). It is positioned just before the LLM \( f(\cdot) \). During inference, we compute the importance score for each visual token with the classifier as follows:
\begin{equation}
\label{eq:computeS}
\begin{aligned}
    \mathbb{S} = p_{\thetav}( \mathbf{H}_\texttt{v} , \mathbf{H}_t )
\end{aligned},
\end{equation}
where \( \mathbb{S} \) represents the importance scores. Based on these scores, we perform pruning with a given reserve ratio \( r \). The indices of the visual tokens to be retained are determined by:
\begin{equation}
\label{eq:computeI}
    \mathbb{I} = \text{TopK}(\mathbb{S}, r \times n_v) ,
\end{equation}
where \( n_v \) is the total number of visual tokens.

Note that \( \mathbb{S} \) is computed over visual and text tokens, but only the scores for visual tokens. This is because visual token redundancy depends not only on visual features but also on textual context, which guides the model in identifying more relevant visual tokens~\cite{sun2025lvpruning}.

\subsection{Training Stage 1}
Our observations indicate that attention weights in deeper layers of the LLM effectively guide the pruning process. Attention weights in these layers contain significant contextual information, which highlights the tokens that need to be attended by the attention mechanism. Therefore, we leverage the information to train the classifier as in Figure~\ref{fig:Train1}.

We denote the guiding attention weights as \( A_g \), and the specific layer from which these weights are derived as \( l_g \). For training, we use the accumulated attention weights, denoted as \( A_l \), as the target labels. The label for the \( k \)-th token is computed by accumulating the attention weights over the relevant layers:
\begin{equation}
\label{eq:accatt}
    A_{l, k} = \sum_{i=1}^{h}\sum_{j=n-n_v}^{n} A_{g, i, j,k} ,
\end{equation}
where \( h \) represents the number of attention heads, \( n \) is the number of tokens.

The classifier outputs \( \hat{\mathbb{S}} \) represents the predicted importance scores for the visual tokens. To train the classifier, we optimize the model using the mean squared error (MSE) loss function, aiming to minimize the discrepancy between the predicted scores \( \hat{\mathbb{S}} \) and the accumulated attention labels \( A_l \). The MSE loss is computed as follows:

\begin{equation}
    \mathcal{L}_{\text{mse}} = \frac{1}{n_v} \sum_{k=1}^{n_v} \left( \hat{\mathbb{S}}_k - A_{l,k} \right)^2,
\end{equation}
where \( \hat{\mathbb{S}}_k \) , \( A_{l,k} \) are the predicted and true importance scores, respectively, for the \( k \)-th visual token. 

The goal is to train the classifier to output scores that align with the accumulated attention weights, which will guide the pruning operation effectively.

This training is efficient, as only a small classifier \( p_{\thetav}(\cdot) \) is optimized. The LLM parameters before layer \( l_g \) remain fixed, avoiding gradient computation, while those after layer \( l_g \) are dropped, greatly reducing both computation and memory overhead.


\subsection{Training Stage 2}
To further improve the model's capacity to capture contextual information and accurately identify important visual tokens, we introduce an end-to-end training phase as Figure~\ref{fig:Train2}, incorporating a differentiable approximation of the pruning operation.

Direct pruning of less-relevant visual tokens during training using hard indexing (e.g., \( \mathbf{H}_\texttt{v}[ \mathbb{I} ] \)) is non-differentiable and thus breaks the backpropagation process. To address this, we simulate the pruning effect by modifying the attention mechanism through a soft attention mask.

We apply a sigmoid activation to the classifier outputs \( \hat{\mathbb{S}} \) to normalize the predicted importance scores into the range \([0, 1]\):
\begin{equation}
    \mathbb{P} = \sigma(\hat{\mathbb{S}}).
\end{equation}
Here, \( \mathbb{P}_i \) can be interpreted as the retention probability for the \( i \)-th visual token. To simulate pruning, we convert the normalized importance scores into attention biases using a logarithmic transformation:
\begin{equation}
    B = \log(\mathbb{P}).
\end{equation}
This transformation ensures that tokens with low importance scores receive large negative biases, thus masking them during attention computation.

We then construct the final attention mask in Figure~\ref{fig:attn_mask} by adding the attention bias \( B_j \) to the standard causal attention mask:
\begin{equation}
    M_{i,j} = M_{i,j}^{\text{causal}} + B_j, \quad \text{for } i > j,
\end{equation}
where \( M^{\text{causal}} \) is the standard causal mask. 

The model outputs predictions \( \hat{y} \), and the ground truth labels are denoted as \( y \). We define the training objective as a combination of the cross-entropy loss and a regularization term:
\begin{equation}
    \mathcal{L}_{\text{total}} = \mathcal{L}_{\text{ce}}(y, \hat{y}) + \mathit{k} \times \mathcal{L}_{\text{reg}}(\mathbb{P}),
    \label{eq:loss_2}
\end{equation}
where \( \mathcal{L}_{\text{reg}}(\mathbb{P}) \) enforces the model to retain a predefined ratio \( r \) of visual tokens and encourages the model to approximate the token pruning patterns during inference. The parameter $\mathit{k}$ controls the weight of \( \mathcal{L}_{\text{reg}}(\mathbb{P}) \) in the overall loss function.

A naive regularization \cite{DMC} such as:$  \mathcal{L}_{\text{reg}} = \mathcal{L}_1(r, \text{mean}(\mathbb{P})) $ enforces a global retention rate \( r \), but tends to collapse all probabilities \( \mathbb{P}_i \) to values near \( r \), harming discriminative capacity.

To promote a clearer distinction between important and unimportant visual tokens, we introduce a contrastive style regularization objective that explicitly separates their predicted importance scores. 

\begin{table*}[t]
\centering
\renewcommand{\arraystretch}{0.9}
\resizebox{\textwidth}{!}{
\begin{tabular}{l cccccccc|c}
\toprule
Models & GQA & MME & SEED & MMStar & AI2D & OCRVQA & TextVQA & InfoVQA & Avg(\%) \\
\midrule
LLaVA-OV-7b   & 61.70 & 1605.41 & 76.59 & 61.67 & 82.77 & 59.83 & 75.02 & 65.52 & 100.00\% \\
\midrule
\multicolumn{10}{c}{reserve ratio = 0.5} \\
\midrule
FastV         & 60.89 & 1586.18 & 74.95 & 57.20 & 80.70 & \underline{58.56} & 71.69 & 49.58 & 94.34\% \\
SparseVLM     & 59.35 & 1560.75 & 74.40 & 54.67 & 78.69 & 45.96 & 69.49 & 42.06 & 88.49\% \\
PDrop         & \underline{61.02} & 1590.56 & \underline{75.75} & \underline{59.00} & 80.83 & 57.19 & \underline{74.71} & \textbf{60.90} & \underline{97.29\%} \\
PDrop*        & 59.97 & 1532.82 & \textbf{75.89} & 58.00 & 80.24 & \textbf{59.90} & 70.64 & 46.83 & 93.56\% \\
CoViPAL       & \textbf{61.31} & \textbf{1613.37} & 75.48 & \textbf{59.07} & \textbf{82.12} & 57.85 & \textbf{74.08} & \underline{59.66} & \textbf{97.48\%} \\
\midrule
\multicolumn{10}{c}{reserve ratio = 0.25} \\
\midrule
FastV         & 56.12 & 1523.37 & 65.88 & 47.23 & 73.25 & 46.29 & 57.29 & 35.18 & 80.54\% \\
SparseVLM     & 52.85 & 1415.58 & 67.42 & 45.40 & 70.30 & 32.09 & 43.61 & 28.03 & 71.88\% \\
PDrop         & \underline{58.02} & 1470.22 & 67.50 & \underline{49.80} & 73.06 & 48.83 & \textbf{68.32} & \underline{41.90} & \underline{84.93\%} \\
PDrop*        & 57.77 & \underline{1531.10} & \underline{70.47} & 49.80 & \underline{74.31} & \textbf{49.22} & 64.95 & 34.47 & 84.12\% \\
CoViPAL       & \textbf{59.93} & \textbf{1559.29} & \textbf{73.22} & \textbf{54.33} & \textbf{79.47} & \underline{48.92} & \underline{65.99} & \textbf{47.28} & \textbf{89.48\%} \\
\bottomrule
\end{tabular}
}
\caption{Image benchmark results. PDrop and PDrop* represent the training-free and training-based versions of PyramidDrop, respectively, which is consistent in the subsequent tables.}
\label{tb:imageresults}
\vspace{-3mm}
\end{table*}

\begin{table*}[t]
\centering
\renewcommand{\arraystretch}{0.9}
\resizebox{\textwidth}{!}{
\begin{tabular}{l| cccccc |c}
\toprule
Models & MVBench & MMBVideo & $\text{MLVU}^{m}$ & $\text{MLVU}^{g}$ & LongVB & WorldSense & Avg(\%) \\
\midrule
LLaVA-Video-7b & 58.32 & 1.71 & 62.40 & 4.16 & 52.50 & 38.20 & 100.00\% \\
\midrule
\multicolumn{8}{c}{reserve ratio = 0.5} \\
\midrule
FastV & \textbf{56.87} & \textbf{1.67} & 60.60 & 4.89 & \textbf{52.60} & 37.60 & \underline{101.41\%} \\
SparseVLM & 55.29 & 1.63 & 59.20 & 4.51 & 50.10 & 37.30 & 97.74\% \\
PDrop & 55.21 & 1.63 & 56.80 & \underline{4.96} & \underline{52.10} & 34.80 & 98.43\% \\
PDrop* & 55.74 & 1.60 & \textbf{61.50} & 4.73 & 49.60 & \textbf{38.90} & 99.62\% \\
CoViPAL & \underline{56.66} & \underline{1.66} & \underline{61.40} & \textbf{4.97} & 51.80 & \underline{38.10} & \textbf{101.75\%} \\
\midrule
\multicolumn{8}{c}{reserve ratio = 0.25} \\
\midrule
FastV & 52.74 & 1.55 & \underline{55.90} & 4.68 & 48.20 & 36.50 & 95.08\% \\
SparseVLM & 50.00 & 1.52 & 54.50 & 4.33 & 47.00 & 36.30 & 91.77\% \\
PDrop & 50.50 & 1.55 & 53.30 & 4.70 & \underline{48.90} & 33.50 & 92.74\% \\
PDrop* & \underline{53.03} & \underline{1.58} & \textbf{59.20} & \underline{4.72} & 48.30 & \textbf{37.70} & \underline{97.06\%} \\
CoViPAL & \textbf{55.42} & \textbf{1.61} & 55.80 & \textbf{4.85} & \textbf{51.30} & \underline{37.20} & \textbf{98.38\%} \\
\bottomrule
\end{tabular}
}
\caption{Results of video benchmarks.}
\label{tb:videoresult}
\vspace{-3mm}
\end{table*}

We first compute the indices of the top and bottom tokens based on the classifier's normalized outputs \( \mathbb{P} \in [0,1]^{n_v} \), where \( n_v \) is the number of visual tokens:
\begin{equation}
\begin{aligned}
    \mathbb{I}_{\text{high}} &= \text{TopK}(\mathbb{P}, \lfloor r \cdot n_v \rfloor), \\
    \mathbb{P}_{\text{high}} &= \mathbb{P}[\mathbb{I}_{\text{high}}], \\
    \mathbb{I}_{\text{low}} &= \text{DTopK}(\mathbb{P}, \lfloor (1 - r) \cdot n_v \rfloor), \\
    \mathbb{P}_{\text{low}} &= \mathbb{P}[\mathbb{I}_{\text{low}}].
\end{aligned}
\end{equation}

Then we define the regularization loss as:
\begin{equation}
    \mathcal{L}_{\text{reg}} = \mathcal{L}_1\left(1, \text{mean}(\mathbb{P}_{\text{high}}) - \text{mean}(\mathbb{P}_{\text{low}})\right).
\end{equation}
This objective aims to maximize the average margin between the most and least important tokens. Specifically:
\( \text{TopK}(\cdot) \) returns the indices of the top \( r \cdot n_v \) visual tokens with the highest importance scores,
 \( \text{DTopK}(\cdot) \) returns the indices of the bottom \( (1 - r) \cdot n_v \) tokens,
 \( \mathcal{L}_1(1, \cdot) \) penalizes deviation from the target margin of 1 between high and low importance scores.

This regularization guides the classifier to assign high retention scores to top-ranked tokens and low scores to less relevant ones, aligning with the inference-time selection and enabling pruning-aware learning in a fully differentiable way.

\section{Experiments}
\label{seq:experiments}
\subsection{Experimental Setup}
\paragraph{Baselines} We evaluate our methods with three baseline approaches: FastV~\cite{chen2024image}, SparseVLM~\cite{zhang2024sparsevlm}, and PyramidDrop~\cite{PyramidDrop}, all of which performing token pruning. FastV prunes visual tokens in a specific layer using self-attention scores of that layer. PyramidDrop prunes tokens in predefined layers based on attention weights. SparseVLM also prunes tokens in predefined layers but merges part of the pruned tokens and reserve them. FastV and SparseVLM are plug-and-play methods, while PyramidDrop offers both training-free and training-based strategies.

\paragraph{Base Models} We conduct experiments on two state-of-the-art LVLMs: LLaVA-OneVision-7b-Chat~\cite{xiong2024llavaovchat} and LLaVA-Video-7b~\cite{llava-vedio}. LLaVA-OneVision-7b-Chat is trained on a combination 4.8M dataset of image and video. LLaVA-Video-7b is fine-tuned from LLaVA-OneVision using a joint dataset, including LLaVA-Video-178K. For evaluating image tasks, we use LLaVA-OneVision-7b-Chat, while LLaVA-Video-7b is used for video tasks.

\paragraph{Classifier Model} We design a compact classifier with two projection layers and 8 encoder layers. The first projection maps LVLM embeddings to the classifier input, and the second outputs a scalar score \( \mathbb{S} = 1 \). The encoder comprises 8 layers, each with a hidden size of 768, intermediate size of 3072, 16 attention heads, and 4 key-value heads, resulting in a total of 71.20M parameters.

\paragraph{Training Implementation} In training stage 1, we use 3\% of LLaVA-NeXT-Data (which is 0.46\% of the training data of LLaVA-OneVision-7b-Chat), totaling 22.2K samples, to train the classifier with base model LLaVA-OneVision-7b-Chat. After stage 1, we proceed to Stage 2, initializing the classifier from Stage 1. During the training stage 2, we trained two classifiers. One is trained on the same data as training stage 1 with the base model LLaVA-OneVision-7b-Chat, which is used for image benchmark evaluation. And another on 20\% of the 0\_30\_s\_academic\_v0\_1 (13.2K samples) dataset with LLaVA-Video-7b for video benchmark. For PyramidDrop, we fine-tune two models with LoRA~\cite{hu2022lora}: one on 10\% of LLaVA-NeXT-Data with LLaVA-OneVision-7b-Chat for image evaluation, and the other on 60\% of 0\_30\_s\_academic\_v0\_1 with LLaVA-Video-7b for video benchmark. The larger dataset for PyramidDrop ensures consistent training time, as Stage 2 is incompatible with Flash Attention~\cite{dao2022flashattention}, which doesn't support this type of custom attention mask currently. Detailed compatibility of Flash Attention is available in Appendix~\ref{sec:app_flash_attention}.

\begin{figure*}[htbp]
    \centering
    \includegraphics[width=0.9\linewidth]{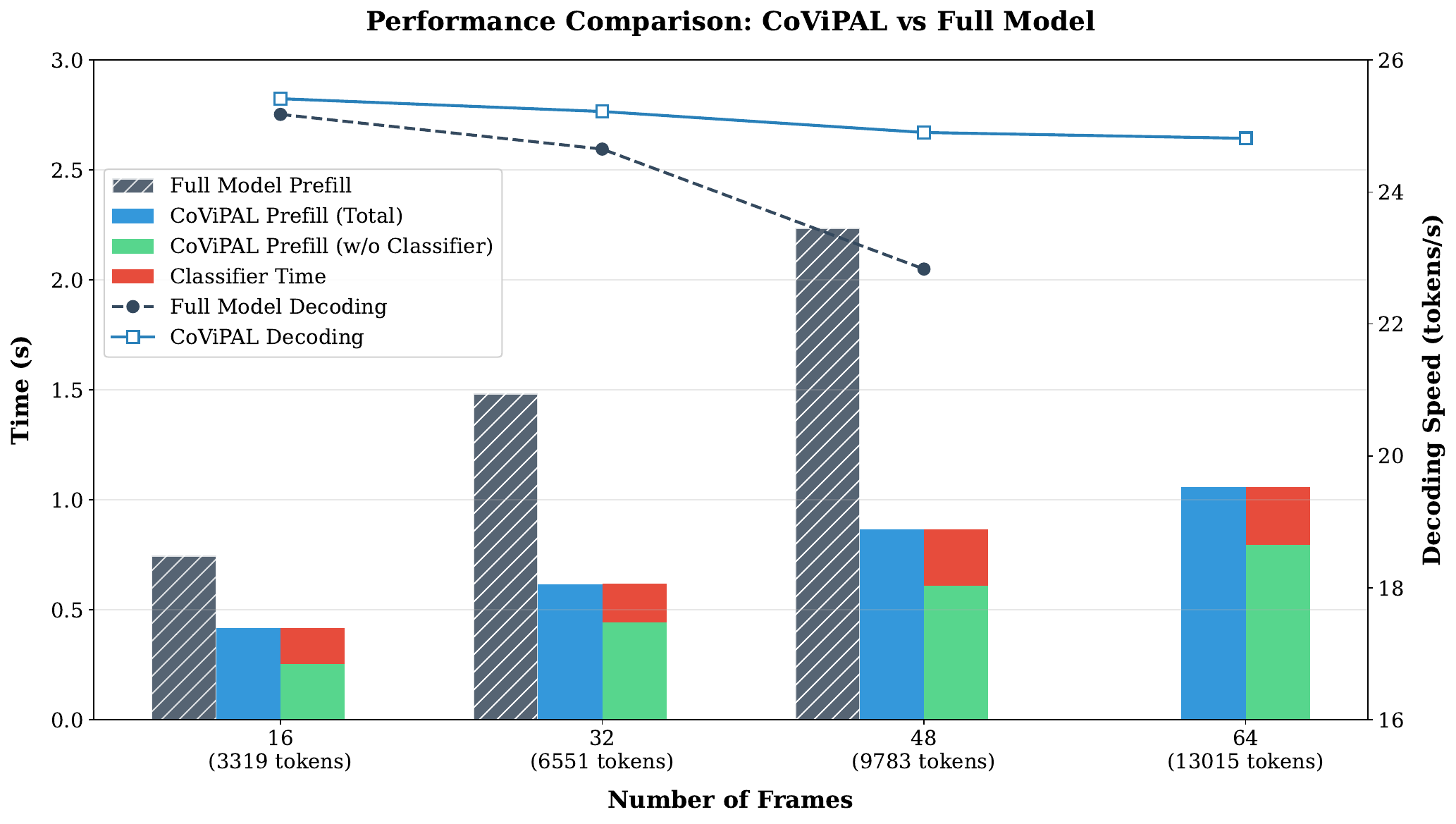}
    \caption{Efficiency Results of CoViPAL on LLaVA-OneVision-7b-Chat.}
    \label{fig:prefill_decoding}
\end{figure*}

\paragraph{Training Hyperparameters}  
Each run is trained for one epoch using Bfloat16 precision. The learning rate is set to 1e-5, except in Stage 2 where it is reduced to 0.5e-5 to preserve parameters in Stage 1. We set \( \mathit{k} = 0.01 \) in Eq.~\ref{eq:loss_2}, and apply a cosine scheduler. For PyramidDrop, we use a LoRA rank of 32 (97.72M trainable params). The input length is capped at 3000 tokens. For image input, we use the anyres-max-2 setting, producing up to 2189 visual tokens-leaving room for text to avoid truncation. For video input, we allow 8 frames (max 1568 visual tokens). The reserve ratio is fixed at 0.25, as we believe that a training-based method should be robust enough to accommodate differences in reserve ratio between training and inference.

\paragraph{Evaluation Benchmarks}  
We evaluate our methods on eight image and five video benchmarks spanning visual reasoning, multimodal comprehension, temporal understanding and so on. This diverse set ensures a comprehensive assessment across visual inputs. Details are provided in the Appendix~\ref{app:benchmark_detail}. All evaluations use VLMEvalKit~\cite{duan2024vlmevalkit}.

\begin{table}[t]
\centering
\renewcommand{\arraystretch}{0.9}
\resizebox{\linewidth}{!}{ 
\begin{tabular}{l cccc}
\toprule
Method & 16Frame & 32Frame &48Frame &64Frame   \\
\midrule
\multicolumn{5}{c}{Prefilling overhead (ms, $\downarrow$)} \\
\midrule
FullKV     &745.0 &1482.1 &2236.2 &OOM  \\
FastV      &\textbf{340.4} &\textbf{513.8}  &1056.0 &OOM \\
SparseVLM  &464.6 &677.4  &1494.3 &OOM  \\
PDrop      &433.1 &\underline{602.5}  &\textbf{795.9}  &OOM \\
CoViPAL    &\underline{416.4} &616.8  &\underline{865.2}  &\textbf{1058.3}  \\
\midrule
\multicolumn{5}{c}{Decoding speed (tokens/second, $\uparrow$)} \\
\midrule
FullKV     &\underline{25.18} &\underline{24.65} &22.83 &OOM  \\
FastV      &24.64 &24.55 &23.10 &OOM \\
SparseVLM  &24.17 &24.23 &24.10 &OOM  \\
PDrop      &24.45 &24.61 &\underline{24.18} &OOM \\
CoViPAL    &\textbf{25.41} &\textbf{25.22} &\textbf{24.90} &\textbf{24.81}  \\
\bottomrule
\end{tabular}
}
\caption{Prefilling overhead and decoding speed of CoViPAL and baselines.}
\label{tb:prefill_decoding}
\vspace{-2mm}
\end{table}

\subsection{Evaluation Results}
\paragraph{Image Benchmarks}
We evaluate CoViPAL on eight widely used image benchmarks, with the results reported in Table \ref{tb:imageresults}. Our results indicate that CoViPAL effectively preserves the model's image comprehension capabilities on tasks of real-world scenarios. CoViPAL consistently surpasses the three baseline methods when retaining 50\% or only 25\% of the image tokens. Particularly, when the reserve ratio is set to 25\%, which significantly challenges the model's token selection capability, CoViPAL demonstrates superior performance by accurately identifying and preserving the most crucial visual tokens. Additionally, results confirm the robustness of CoViPAL, as performance remains stable even when the inference reserve ratio (50\%) differs from the training reserve ratio (25\%).


\begin{table}[t]
\centering
\renewcommand{\arraystretch}{1.2}
\resizebox{0.95\linewidth}{!}{ 
\begin{tabular}{l ccc}
\toprule
Reserve Ratio & Classifier & Prefilling & Decoding  \\
\midrule
FullKV 100\%     & - & 17679 & 16817 \\
CoViPAL 90\%     & 16323 & 17804 & 16795 \\
CoViPAL 75\%     & 16323 & 17534 & 16699 \\
CoViPAL 50\%     & 16323 & 16723 & 16539 \\
CoViPAL 25\%     & 16323 & 16639 & 16377 \\
CoViPAL 10\%     & 16323 & 16378 & 16281 \\
\bottomrule
\end{tabular}
}
\caption{Memory consumption of CoViPAL.}
\label{tb:memory_consumption}
\vspace{-2mm}
\end{table}

\begin{table*}[t]
\centering
\renewcommand{\arraystretch}{0.9}
\resizebox{0.95\textwidth}{!}{
\begin{tabular}{l cccccccc}
\toprule
Models & GQA & MME & SEED & MMStar & AI2D & OCRVQA & TextVQA & InfoVQA  \\
\midrule
LLaVA-OV-7b   & 61.70 & 1605.41 & 76.59 & 61.67 & 82.77 & 59.83 & 75.02 & 65.52\\
\midrule
\multicolumn{9}{c}{reserve ratio = 0.5} \\
\midrule
$ p_{\thetav}^1$     &60.85 &1547.10 &74.85 &58.20 &81.22 &52.77 &68.35 &52.12  \\
$ p_{\thetav}^8$     &\textbf{61.31} &\textbf{1613.37} &\textbf{75.48} &\textbf{59.07} &\textbf{82.12} &\textbf{57.85} &\textbf{74.08} &\textbf{59.66} \\
\midrule
\multicolumn{9}{c}{reserve ratio = 0.25} \\
\midrule
$ p_{\thetav}^1$     &57.21 &1446.89 &70.67 &51.20 &76.91 &37.43 &48.21 &35.55  \\
$ p_{\thetav}^8$     &\textbf{59.93} &\textbf{1559.29} &\textbf{73.22} &\textbf{54.33} &\textbf{79.47} &\textbf{48.92} &\textbf{65.99}  &\textbf{47.28} \\
\bottomrule
\end{tabular}
}
\caption{Performance comparison of image benchmarks on two model architectures.}
\label{tb:ablation_model}
\vspace{-3mm}
\end{table*}

\paragraph{Video Benchmarks}
We further evaluate CoViPAL on five widely recognized video benchmarks, with the results summarized in Table \ref{tb:videoresult}. The experimental results demonstrate that CoViPAL effectively eliminates redundant or less relevant visual tokens, leading to performance improvements under various conditions. CoViPAL consistently outperforms the three comparative baselines, exhibiting only a minor performance degradation of 1.62\% when pruning 75\% of the visual tokens. Moreover, the results suggest that videos are more information-sparse compared to images, containing a higher proportion of redundant visual tokens, thereby making video tasks inherently more robust to token pruning.



\begin{table}[t]
\centering
\renewcommand{\arraystretch}{1.0}
\begin{subtable}{0.44\linewidth}
\centering
\resizebox{!}{1.23cm}{
\begin{tabular}{l cc}
\toprule
\multirow{2}{*}{Depth} & \multicolumn{2}{c}{Reserve Ratio} \\
\cmidrule(lr){2-3}
& 0.5 & 0.25 \\
\midrule
2 & 61.00 & 57.75 \\
4 & 61.05 & 58.53 \\
8 & 61.31 & 59.93 \\
\bottomrule
\end{tabular}
}
\caption{Different model depth.}
\label{tb:classifier_depth}
\end{subtable}
\hfill
\begin{subtable}{0.54\linewidth}
\centering
\resizebox{!}{1.23cm}{
\begin{tabular}{l cc}
\toprule
\multirow{2}{*}{Hidden Size} & \multicolumn{2}{c}{Reserve Ratio} \\
\cmidrule(lr){2-3}
& 0.5 & 0.25 \\
\midrule
384 & 61.42 & 57.75 \\
768 & 61.31 & 59.93 \\
1536 & 60.91 & 59.96 \\
\bottomrule
\end{tabular}
}
\caption{Different model width.}
\label{tb:classifier_width}
\end{subtable}
\caption{GQA benchmark results for different model depth and model width.}
\label{tb:classifier_ablation}
\vspace{-2mm}
\end{table}

\paragraph{Efficiency Results}
We evaluate the efficiency of CoViPAL on LLaVA-OneVision-7b-Chat with video input on a single 
RTX 3090 24G GPU. The sample frame size ranges from 16 to 64, resulting in input tokens ranging from 3k to 13k. With a reserve ratio of 0.25, we measure prefilling time, decoding speed and memory consumption for generating 1k tokens, and the classifier's overhead during prefilling. All methods are integrated with FlashAttention. Results are shown in Figure~\ref{fig:prefill_decoding}, Table~\ref{tb:prefill_decoding} and Table~\ref{tb:memory_consumption}.

Our method substantially reduces prefilling time and accelerates decoding. In terms of decoding speed, our method consistently outperforms all baseline approaches. For the prefilling stage, the introduction of the classifier model incurs negligible time overhead, and the overall performance of our method is comparable to the baselines. Notably, for 48-frame inputs, CoViPAL reduces prefilling time by over 60\% and enables 64-frame inference on a 24GB GPU, whereas the original model and all baselines fail due to memory limitations.

CoViPAL consistently reduces the decoding peak memory, achieving over 1 GiB memory savings when pruning 75\% of tokens. This demonstrates CoViPAL' s promise for high-throughput LVLM applications.

\subsection{Ablation Study}  

\paragraph{Model Structure for Contextual Information Capture} 
We first compare two classifier models: a multi-layer encoder with 71.2M parameters (\( p_{\thetav}^8 \)) and a single-layer encoder with 165.18M parameters (\( p_{\thetav}^1 \)), in which the latter uses the same settings as LLaVA-OneVision-7b-Chat decoder.

Trained with the same two-stage strategy on 3\% of LLaVA-NeXT-Data, \( p_{\thetav}^8 \) consistently outperforms \( p_{\thetav}^1 \) on image tasks, as shown in Table~\ref{tb:ablation_model}. Despite its smaller size, the deeper model captures redundant token patterns more effectively, highlighting the advantage of deeper attention layers in modeling contextual information for pruning.

Subsequently, we evaluate the impact of varying model depth and hidden size on the GQA benchmark. As shown in Table~\ref{tb:classifier_ablation}, more encoder layers or a larger hidden size help the classifier better capture visual features and classify redundant tokens. The classifier architecture with different hyperparameters shows consistent performance trends, which proves the generality of our method.

\begin{table}[t]
\centering
\renewcommand{\arraystretch}{0.9}
\resizebox{0.8\linewidth}{!}{ 
\begin{tabular}{l cc}
\toprule
Models & $r=50\%$ & $r=25\%$  \\
\midrule
LLaVA-OV-7b   &\multicolumn{2}{c}{61.70} \\
\midrule
$\mathit{k}=0.1 $     &61.19 &59.31  \\
$\mathit{k}=0.01 $      &\textbf{61.31} &\textbf{59.94} \\
$\mathit{k}=0.0001 $      &61.11 &58.73  \\
\bottomrule
\end{tabular}
}
\caption{Ablation study on \( \mathit{k} \).}
\label{tb:ablation_k}
\vspace{-2mm}
\end{table}


\paragraph{$\mathit{\textbf{k}}$ for Training Stage 2}  
The hyperparameter \( \mathit{k} \) in Eq.~\ref{eq:loss_2} is crucial in Stage 2. A large \( \mathit{k} \) causes early sharp separation of retain probabilities \( \mathbb{P} \) which hindering the subsequent training, while a small \( \mathit{k} \) keeps \( \mathbb{P} \) continuous, misaligned with the discrete selection required during inference. The distribution of classifier outputs are showed in Appendix~\ref{sec:app_classifier_out}.

We train with \( \mathit{k} \) values from 0.0001 to 0.1 using LLaVA-OneVision-7b-Chat and evaluate on GQA~\cite{hudson2019gqa}. For \( \mathit{k} = 0.0001 \), we warm up with \( \mathit{k} = 0.01 \) to avoid continuous distribution throughout training. As shown in Table~\ref{tb:ablation_k}, \( \mathit{k} = 0.01 \) yields the best performance.



\begin{figure}[t]
    \centering
    \begin{subfigure}[b]{0.43\linewidth}
        \centering
        \includegraphics[width=\linewidth]{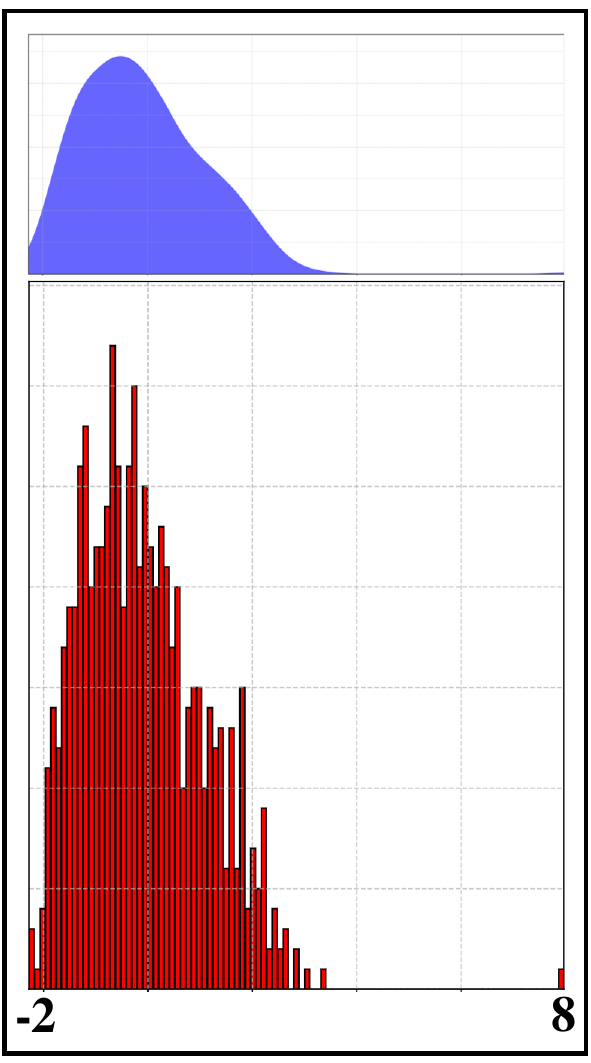}
        \caption{$\mathbb{S}$ of classifier after training stage 1.}
        \label{fig:stage1.pdf}
    \end{subfigure}
    \hspace{0.03\linewidth}
    \begin{subfigure}[b]{0.43\linewidth}
        \centering
        \includegraphics[width=\linewidth]{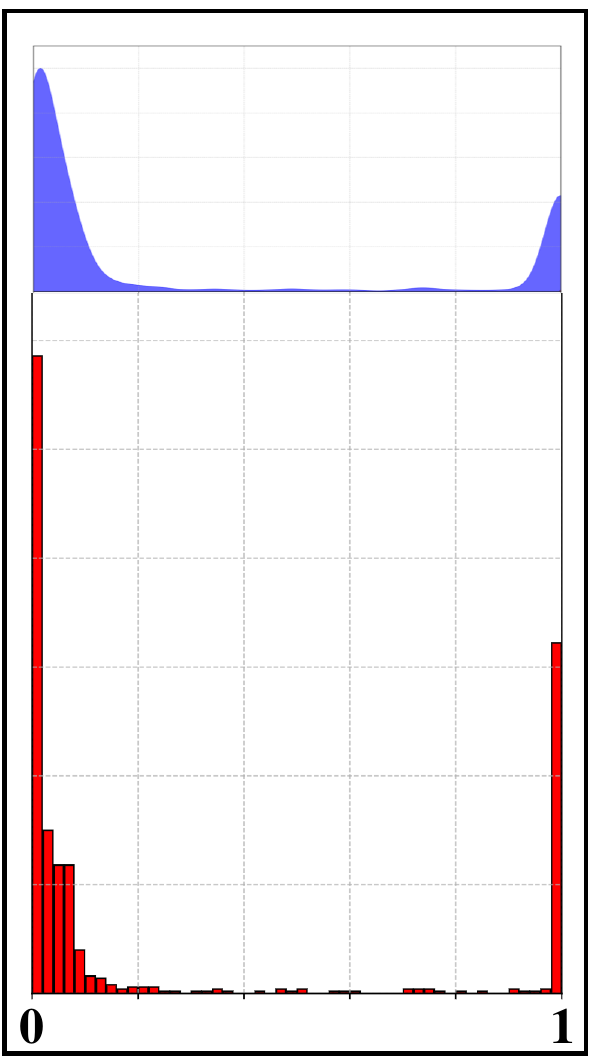}
        \caption{$\mathbb{P}$ of classifier after training stage 2.}
        \label{fig:stage2_0.01.pdf}
    \end{subfigure}
    \caption{The distribution of classifier outputs after two training stages.}
    \label{fig:two_stage_training}
\end{figure}

\paragraph{Effectiveness of the Training Strategy}  
Training Stage 2 from a randomly initialized model led to a collapse of retain probabilities \( \mathbb{P} \) to 0 throughout training, even with \( \mathit{k} = 0.1 \), as shown in Figure~\ref{appfig:stage2_0.0001.pdf}. In contrast, initializing from the Stage 1 model Figure~\ref{fig:stage1.pdf} allowed \( \mathbb{P} \) to stabilize and discretize effectively Figure~\ref{fig:stage2_0.01.pdf}.

These results underscore the value of the two-stage strategy: Stage 1 captures contextual attention patterns, providing a strong initialization for Stage 2 to identify redundant tokens and simulate pruning under smaller \( \mathit{k} \).

More experiment results and analysis are available in Appendix~\ref{sec:app_more_experiments}.

\section{Conclusion}
We propose CoViPAL, a novel contextualized visual token pruning method that efficiently reduces the computational and memory overhead of Large Vision-Language Models by leveraging a lightweight and plug-and-play pruning module. CoViPAL identifies and removes redundant visual tokens across all layers with minimal supervision, achieving up to 50\% reduction in pre-filling time and pruning 75\% of visual tokens while maintaining competitive performance. Our method outperforms both training-free and training-based approaches, offering a scalable and adaptable solution for efficient multimodal inference. This work provides new insights into visual token redundancy and paves the way for deploying LVLMs in resource-constrained settings.

\section*{Limitations}
While our approach has been validated on representative LVLMs, the diversity of model backbones explored so far remains limited. In future work, we plan to extend our method to a broader range of architectures, including base models from the LLaMA and Mistral families, to assess its applicability across different LVLM paradigms and better understand its architectural generality.

In addition, the current experiments are conducted on models of moderate scale. Scaling up to larger model sizes will allow us to further investigate the generalization and effectiveness of our pruning framework in high-capacity settings. These extensions will provide deeper insights into the scalability and robustness of our approach.


\bibliography{custom}
\appendix
\clearpage

\section{Related Work}
\label{sec:appendix}
\subsection{Large Vision-Language Models}
\label{sec:LVLM}
Large vision-language models (LVLMs) combine vision encoders with large language models to jointly process image and text inputs. This multimodal architecture has achieved strong performance on tasks like visual question answering and captioning, with representative models including BLIP-2\cite{li2023blip}, Qwen-VL~\cite{qwenvl}, and the LLaVA series~\cite{liu2024llavanext}. LLaVA-OneVision~\cite{llava-onevision} extends LVLMs to handle single-image, multi-image, and video inputs in a unified framework, while LLaVA-Video~\cite{llava-vedio} adapts to the video domain via instruction tuning. However, rich visual inputs often produce thousands of tokens, leading to high computational and memory costs. This bottleneck limits inference efficiency and practical deployment, highlighting the need for token compression to make LVLMs more scalable and efficient.

\section{Flash Attention Compatibility}
\label{sec:app_flash_attention}
Our method is compatible with Flash Attention during both inference and Stage 1 of training (for both the classifier model and the LVLM). The only incompatibility arises in Stage 2 of LVLM: Flash Attention currently does not support the custom attention mask depicted in Figure~\ref{fig:attn_mask}. Consequently, we fall back to eager (standard) attention for the LVLM in Stage 2, while the classifier model remains Flash-Attention-compatible throughout.

\section{More Experiments and Analysis}
\label{sec:app_more_experiments}



\subsection{Training with Different Guidance Layer}
 we conduct two-stage training with different guidance layers in stage 1 (stage 2 does not need the guidance any more). We evaluate the performance on the GQA benchmark and the results are indicated in Table~\ref{tb:guidance_layer}. Aligned with our preliminary experiment, attention weights in deeper layers include more contextual information and thus can better guide the pruning process.
 
\begin{table}[t]
\centering
\renewcommand{\arraystretch}{0.9}
\resizebox{0.6\linewidth}{!}{
\begin{tabular}{l cc}
\toprule
Guidance  & \multicolumn{2}{c}{Reserve Ratio} \\
\cmidrule(lr){2-3}
Layer & 0.5 & 0.25 \\
\midrule
0 & 61.01 & 59.34 \\
8 & 60.88 & 58.76 \\
16 & 61.31 & 59.93 \\
\bottomrule
\end{tabular}
}
\caption{Training CoViPAL with Different Guidance Layer.}
\label{tb:guidance_layer}
\vspace{-2mm}
\end{table}

\begin{table*}[t]
\centering
\renewcommand{\arraystretch}{0.9}
\resizebox{0.95\textwidth}{!}{
\begin{tabular}{l cccccccc}
\toprule
Models & GQA & MME & SEED & MMStar & AI2D & OCRVQA & TextVQA & InfoVQA  \\
\midrule
LLaVA-OV-7b   & 61.70 & 1605.41 & 76.59 & 61.67 & 82.77 & 59.83 & 75.02 & 65.52\\
\midrule
Prune at Layer8     &58.95 &1572.62 &74.42 &54.45 &78.95 &56.41 &69.18 &40.71  \\
CoViPAL     &\textbf{61.31} &\textbf{1613.37} &\textbf{75.48} &\textbf{59.07} &\textbf{82.12} &\textbf{57.85} &\textbf{74.08} &\textbf{59.66} \\
\bottomrule
\end{tabular}
}
\caption{Performance comparison of image benchmarks on CoViPAL and pruning at the $8-th$ layer of LVLM.}
\label{tb:prune_at_layer8}
\vspace{-3mm}
\end{table*}

\begin{table*}[t]
\centering
\renewcommand{\arraystretch}{0.9}
\resizebox{0.95\textwidth}{!}{
\begin{tabular}{l cccccccc}
\toprule
Models & GQA & MME & SEED & MMStar & AI2D & OCRVQA & TextVQA & InfoVQA  \\
\midrule
LLaVA-OV-7b   & 61.70 & 1605.41 & 76.59 & 61.67 & 82.77 & 59.83 & 75.02 & 65.52\\
\midrule
PDrop*(3\% data) & 56.77 & 1482.83 & 69.33 & \underline{49.80} & 73.41 & \textbf{49.51} & 64.67 & \underline{35.80} \\
PDrop*(10\% data) & \underline{57.77} & \underline{1531.10} & \underline{70.47} & \underline{49.80} & \underline{74.31} & \underline{49.22} & \underline{64.95} & 34.47 \\
CoViPAL & \textbf{59.93} & \textbf{1559.29} & \textbf{73.22} & \textbf{54.33} & \textbf{79.47} & 48.92 & \textbf{65.99} & \textbf{47.28} \\
\bottomrule
\end{tabular}
}
\caption{Performance comparison of image benchmarks on CoViPAL and PyramidDrop with the same training time or the same dataset.}
\label{tb:pdrop_data10}
\vspace{-3mm}
\end{table*}

\subsection{Comparison with Pruning at 8-th Layer}
we use the attention from layer 8 of the LLM to guide the pruning, and layers 8+ use the pruned tokens during training and inference. We fine-tune LLaVA-OneVision-7B with LoRA, with 97.72M trainable parameters (more than 71.20M of the classifier model). Other experimental settings are the same as those used to train our classifier model. The reserve ratio is 0.5 (pruning 70\% of the visual tokens at layer 8) because pruning at layer 8 cannot achieve a total reserve ratio of 0.25.

As shown in Table~\ref{tb:prune_at_layer8}, with the same depth of 8, CoViPAL achieves a better balance between KV cache size and model performance through pruning visual tokens before the LLM, further proving that some visual tokens are inherently redundant and can be pruned safely when guided by appropriate contextual signals.

\subsection{Comparison with PDrop with the Same Dataset}
To fairly compare with PyramidDrop, we need to ensure the same training time or dataset. We chose the former: we provided more data for PyramidDrop to ensure the same training time. Our method outperforms PyramidDrop with only 1/3 of the training data.

We also train PyramidDrop on 3\% image datasets and reserve 25\% visual tokens. Results are reported on Table~\ref{tb:pdrop_data10}. With the same dataset, PyramidDrop performs far worse than our method. However, this comparison is unfair because PyramidDrop consumed less training time.

\subsection{Classifier Output Distribution}
\label{sec:app_classifier_out}

\begin{figure*}[t]
    \centering
    \begin{subfigure}[b]{0.31\linewidth}
        \centering
        \includegraphics[width=\linewidth]{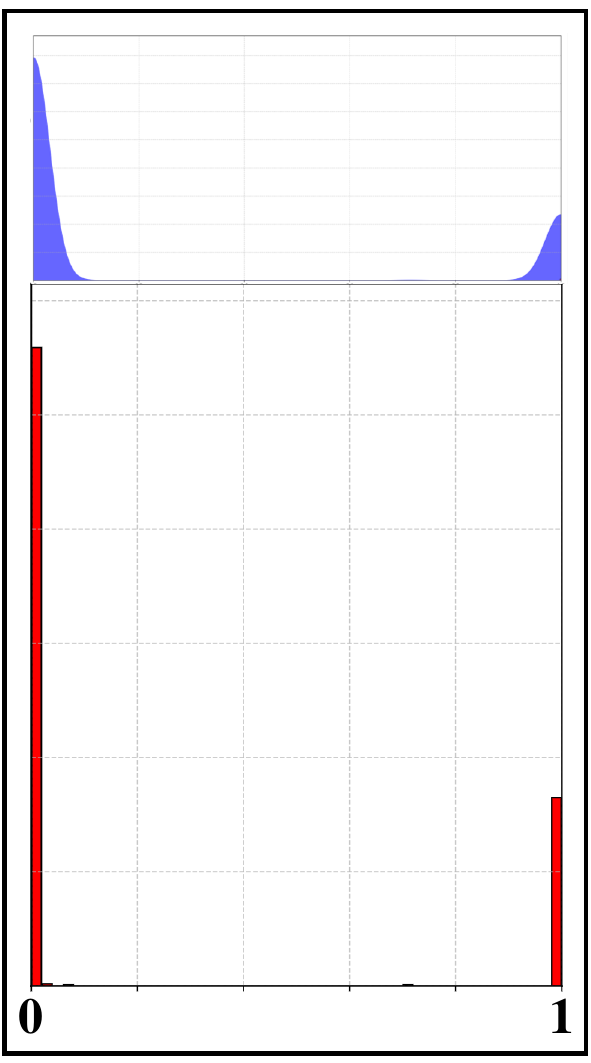}
        \caption{$\mathbb{P}$ of classifier after training stage 2 when $\mathit{k}=0.1$.}
        \label{appfig:stage2_0.1.pdf}
    \end{subfigure}
    \hfill
    \centering
    \begin{subfigure}[b]{0.31\linewidth}
        \centering
        \includegraphics[width=\linewidth]{figures/stage2_0.01.pdf}
        \caption{$\mathbb{P}$ of classifier after training stage 2 when $\mathit{k}=0.01$.}
        \label{appfig:stage2_0.01_app.pdf}
    \end{subfigure}
    \hfill
    \centering
    \begin{subfigure}[b]{0.31\linewidth}
        \centering
        \includegraphics[width=\linewidth]{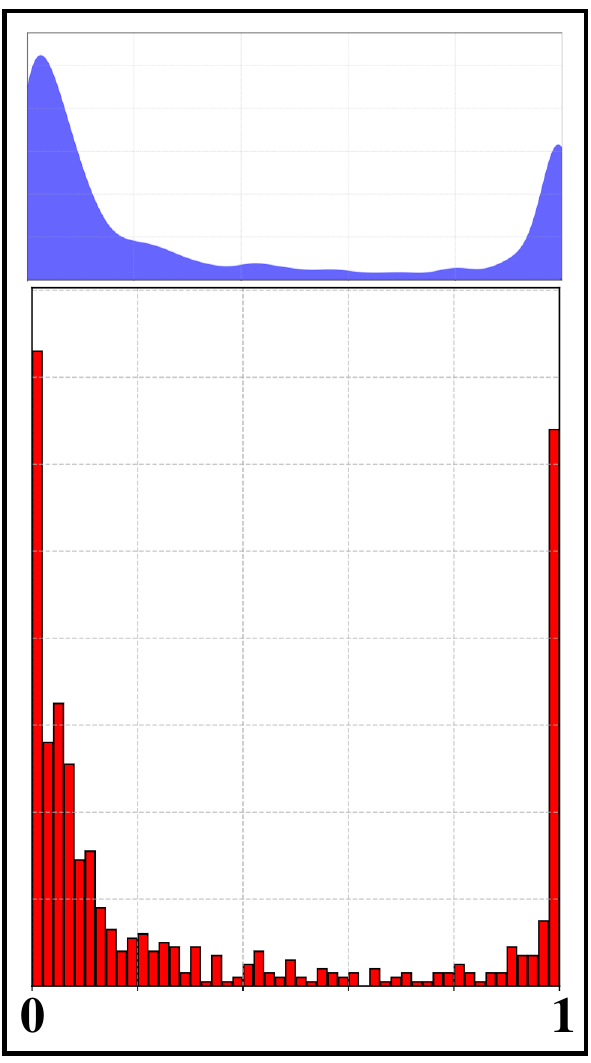}
        \caption{$\mathbb{P}$ of classifier after training stage 2,$\mathit{k}=0.0001$.}
        \label{appfig:stage2_0.0001.pdf}
    \end{subfigure}
    \caption{The distribution of classifier outputs after training stages 2 when setting different $\mathit{k}$.}
    \label{appfig:different_k}
\end{figure*}

We provide the distribution of the classifier model outputs after training stage 2, which highlights the influence of different settings for the hyperparameter \( \mathit{k} \) during this stage. The hyperparameter \( \mathit{k} \) plays a crucial role in the second stage of training. When \( \mathit{k} \) is set to 0.1, the retain probabilities \( \mathbb{P} \) become sharply separated at the beginning of stage 2, as shown in Figure \ref{appfig:stage2_0.1.pdf}, which can hinder subsequent training. On the other hand, when \( \mathit{k} \) is set to 0.0001, a large portion of the \( \mathbb{P} \) values remain continuous, as seen in Figure \ref{appfig:stage2_0.0001.pdf}, which prevents the values from approximating the discrete selection patterns needed during inference. When \( \mathit{k} = 0.1 \), the distribution of the classifier's output in Figure \ref{appfig:stage2_0.01_app.pdf} aligns with the pruning operation during the inference stage, allowing the model to gradually identify redundant tokens and simulate pruning under smaller values of \( \mathit{k} \).

\section{Benchmark Detail}
\label{app:benchmark_detail}
We evaluate our method on a diverse collection of vision-language benchmarks, covering both image and video modalities. As summarized in Table~\autoref{tb:benchmark_detail}, the image-based benchmarks include GQA~\cite{hudson2019gqa}, MME~\cite{fu2024mmecomprehensiveevaluationbenchmark}, SEED-Bench~\cite{li2024seed2plus}, MMStar~\cite{chen2024we}, AI2D~\cite{seo2014diagram}, OCR-VQA~\cite{mishra2019ocr}, TextVQA~\cite{singh2019towards}, and InfographicVQA~\cite{mathew2022infographicvqa}. 

For video-based evaluation, we adopt MVBench~\cite{li2024mvbench}, MMBench-Video~\cite{fang2024mmbench}, MLVU~\cite{zhou2025mlvubenchmarkingmultitasklong}, LongVideoBench~\cite{wu2024longvideobench}, and WorldSense~\cite{hong2025worldsense}. These benchmarks collectively provide a comprehensive testbed for assessing both the effectiveness and generalizability of our proposed method.

\vspace*{-3cm} 

\begin{table*}[t]
    \small
    \centering
    \renewcommand{\arraystretch}{1.3} %
    \resizebox{\textwidth}{!}{
    \begin{tabular}{cccl}
    \toprule
    \textbf{Modality} & \textbf{Benchmark} & \textbf{Short Name} & \textbf{Task Feature} \\
    \midrule
    \multirow{8}{*}{Image} 
    & GQA & GQA & Visual attribute reasoning \\
    & MME & MME & Multimodal evaluation across modalities \\
    & SEED-Bench & SEED & Generative multimodal comprehension \\
    & MMStar & MMStar & Vision tasks with minimal data leakage \\
    & AI2D & AI2D & Diagram understanding \\
    & OCR-VQA & OCRVQA & Text-based image reasoning \\
    & TextVQA & TextVQA & Scene text understanding \\
    & InfographicVQA & InfoVQA & Multimodal infographic reasoning \\
     \midrule
    \multirow{5}{*}{Video} 
    & MVBench & MVBench & Temporal understanding in videos \\
    & MMBench-Video & MMBenchV & Long-form video reasoning \\
    & MLVU & MLVU & Multi-task video understanding \\
    & LongVideoBench & LongVB & Interleaved video-language reasoning \\
    & WorldSense & WorldSense & Omni-modal (visual/audio/text) understanding \\
    \bottomrule
    \end{tabular}
    }
    \caption{Detailed Evaluation Benchmarks }
    \label{tb:benchmark_detail}
    \vspace{-3mm}
\end{table*}


\end{document}